\pdfoutput=1

\documentclass[11pt]{article}

\usepackage[final]{acl}

\usepackage{times}
\usepackage{latexsym}
\usepackage{bm}
\usepackage{amsmath}
\usepackage{algorithm, algpseudocode}
\usepackage[caption=false,font=footnotesize,labelfont=rm,textfont=rm]{subfig}
\usepackage{multirow}
\usepackage{booktabs}
\usepackage{enumitem}
\usepackage{stfloats}
\usepackage{amsfonts}
\usepackage[export]{adjustbox}

\usepackage[T1]{fontenc}

\usepackage[utf8]{inputenc}

\usepackage{microtype}

\usepackage{inconsolata}

\usepackage{graphicx}

\title{DAC: A Dynamic Attention-aware Approach \\for Task-Agnostic Prompt Compression}

\author{Yi Zhao$^{1,2,3}$, Zuchao Li$^{4,*}$ and Hai Zhao$^{1,2,3,}$\thanks{$\ $  Corresponding author ({\tt zcli-charlie@whu.edu.cn and zhaohai@cs.sjtu.edu.cn}). This research was supported by the Joint Research Project of
Yangtze River Delta Science and Technology Innovation Community (No.
2022CSJGG1400), the National Natural Science Foundation of China (No. 62306216), the Natural Science Foundation of Hubei Province of China (No. 2023AFB816) and Xiaomi Open-Competition Research Program.}, Baoyuan Qi$^{5}$, Guoming Liu$^{5}$ \\
$^{1}$School of Computer Science, Shanghai Jiao Tong University\\
$^{2}$Key Laboratory of Shanghai Education Commission for Intelligent Interaction\\ and Cognitive Engineering, Shanghai Jiao Tong University\\
$^{3}$Shanghai Key Laboratory of Trusted Data Circulation and Governance in Web3\\
$^{4}$School of Artificial Intelligence, Wuhan University, Wuhan, China \\
$^{5}$Xiaomi, Beijing, China 
}

\begin{document}
\maketitle
\begin{abstract}
Task-agnostic prompt compression leverages the redundancy in natural language to reduce computational overhead and enhance information density within prompts, especially in long-context scenarios. Existing methods predominantly rely on information entropy as the metric to compress lexical units, aiming to achieve minimal information loss. However, these approaches overlook two critical aspects: (i) the importance of attention-critical tokens at the algorithmic level, and (ii) shifts in information entropy during the compression process. Motivated by these challenges, we propose a dynamic attention-aware approach for task-agnostic prompt compression (DAC). This approach effectively integrates entropy and attention information, dynamically sensing entropy shifts during compression to achieve fine-grained prompt compression. Extensive experiments across various domains, including LongBench, GSM8K, and BBH, show that DAC consistently yields robust and substantial improvements across a diverse range of tasks and LLMs, offering compelling evidence of its efficacy. Our code is available at \href{https://github.com/QQQ-yi/DAC}{https://github.com/QQQ-yi/DAC}
\end{abstract}



\section{Introduction}
Recent advent of In-Context Learning (ICL) \cite{brown2020language,dong2024survey}, Chain-of-Thought (CoT) \cite{wei2022chain,yao2024tree,yao-etal-2024-got, zhang2023ignitinglanguageintelligencehitchhikers}, Retrieval Augmented Generation (RAG) \cite{lewis2020retrieval}, and Autonomous Agent \cite{xi2023rise} technologies has significantly invigorated the landscape of applications based on Large Language Models (LLMs). While these methodologies have expanded the capabilities of LLMs by activating domain-specific knowledge or enhancing memory capacities, they also introduce the challenge of exceedingly long context lengths, which leads to a substantial increase in computation and memory consumption due to inherent self-attention mechanism. 

Efficient LLM is the technology that aims to achieve computational efficiency while retaining performance, which is accomplished through various methods such as modifying model architecture \cite{sun2024you,yang2024kvsharerefficientinferencelayerwise}, parameter quantization \cite{lin2024awq}, key-value (KV) cache compression \cite{yang-etal-2024-pyramidinfer}, and the utilization of soft prompts \cite{mu2024learning}, among others. Despite the effectiveness of these methods, they often require modifications to the model, which is not feasible for black-box LLMs, such as those accessible only through APIs. In such cases, prompt compression, which seeks to shorten the prompt while preserving essential information, represents the most direct way.

Several studies consider essential information to be the parts most relevant to the question or task, thus proposing a task-aware manner for prompt compression \cite{jiang-etal-2024-longllmlingua,xu2023recompimprovingretrievalaugmentedlms,huang-etal-2024-fewer,jung2024discrete}. Benefiting from the sparsity of question-related information in original prompts, these methods achieve significant performance on specific benchmarks with high compression ratios, even surpass the performance of the original prompts in some cases. However, these approaches are highly dependent on the type of downstream task, leading to the following limitations: i) the prompt needs to be repeatedly compressed in scenarios involving multiple questions or tasks, ii) it is challenging to define user’s intent or questions in extended dialogic engagements.

Task-agnostic prompt compression aims to compress prompts relying on the self-information of the language without any additional clues \cite{li-etal-2023-compressing,jiang-etal-2023-llmlingua,pan-etal-2024-llmlingua}. Previous works primarily utilize coarse-grained information entropy output by the logits layer of entire model for compression. They do not delve into the inner layers of LLMs for gathering finer-grained attention scores to enhance the compression process. In addition, these entropy-based methods treat information entropy as static while ignoring the dynamic shift during the compression process. Some low-entropy tokens may become high-entropy tokens after their dependent tokens are compressed.

To address these challenges, we propose a dynamic attention-aware approach for task-agnostic prompt compression (DAC). Specifically, we propose a novel metric for prompt compression that effectively integrates information from both entropy and the attention scores, enabling the compression process to be conducted in an attention-aware manner. Additionally, we propose a dynamic approach to iteratively identify tokens with significant entropy shifts during the compression. This method aims to minimize information loss more precisely, thereby preserving the information density in the compressed prompts.

We evaluate the effectiveness of our approach across different domains, including LongBench for contextual understanding and GSM8K, BBH for reasoning and in-context learning. The experimental results demonstrate that our method significantly outperforms other entropy-based compression methods, achieving an improvement of 4.03 points on LongBench. Moreover, it surpasses current state-of-the-art (SOTA) approaches by an average score of 1.33 on LongBench. Our method also exhibits robust generalization ability, as evidenced by consistent performance improvements across a range of model series, from Qwen2 to LLaMA3.

\section{Related Work}
\subsection{Prompt Compression}

There are two primary approaches in prompt compression, which can be categorized based on the form of the compressed output: soft prompts and hard prompts. The soft prompts methods typically compress the original prompt into non-linguistic forms, such as special tokens or embeddings. \citet{wingate-etal-2022-prompt} proposed optimizing the KL divergence between the answers generated by the original and compressed prompts to achieve soft prompt compression. \citet{mu2024learning} addressed this challenge by introducing GIST tokens. \citet{ge2023in-context} propose ICAE, which pre-trained a compression model to transform prompts into compact memory slots that can be conditioned on by LLM. However, soft prompts often result in non-human-readable formats, leading to difficulties in interpretability of compressed content. Otherwise, these methods frequently require modifications of the model (pre-training or fine-tuning), making them unsuitable for black-box LLMs.

\begin{figure*}[!htb]
\hspace{-10pt}
\includegraphics[scale=0.43]{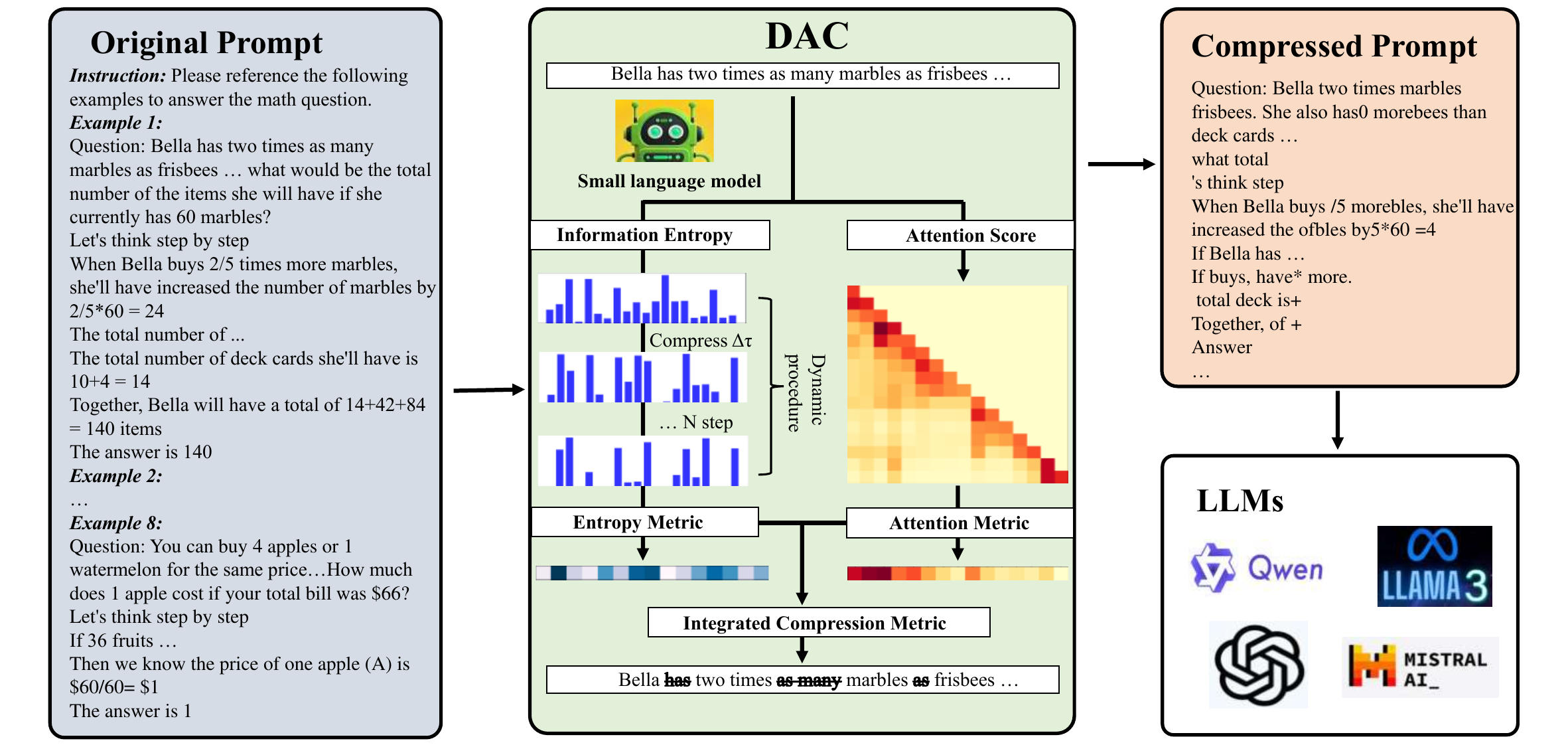}
\caption{Framework of our proposed DAC for task-agnostic prompt compression.}
\label{fig:illus}
\end{figure*}

The hard prompts methods achieve compression by identifying and dropping low-information content from the original prompt. \citet{li-etal-2023-compressing} first introduced using information entropy to measure the information content. They also considered the most effective lexical units for compression and determining using phrases through experiments. Similarly, LLMLingua \cite{jiang-etal-2023-llmlingua} employs information entropy as the metric, proposing the use of a budget controller to assign different compression rates to various parts of the prompt (instruction, demonstrations, and the question). Additionally, they introduced iterative token-level prompt compression, which segments the original prompt and then performs fine-grained token-level compression. These entropy-based methods effectively identify and compress low-information content in the original prompt, achieving satisfactory results in downstream tasks. However, they do not fully account for the information contained in the attention mechanism and the entropy shifts occurring during the compression process. In contrast, LLMLingua2 \cite{pan-etal-2024-llmlingua} takes a different approach by training a specialized classification model dedicated to prompt compression. The compression data used for training is distillated from a more powerful LLM (i.e. GPT4). However, powerful LLMs may not good at compression task, and training on a specific dataset may not generalize well to other tasks. 

Our proposed DAC method falls into this category as well and more simular to entropy-based methods. Differently, DAC employs a novel metric that integrates information from both information entropy and the attention mechanism, and further introduces a dynamic approach for accurately identify low-information content. 


\subsection{KV Cache Compression}
Another line of related work is KV cache compression \cite{shi2024costdownreviewmethods}, as we consider attention mechanism in DAC method. Many previous studies about KV cache compression have analyzed the characteristics in attention matrices of LLM. $H_2O$ \cite{NEURIPS2023_6ceefa7b} observed that the accumulated attention scores of all tokens follow a power-law distribution, indicating that only a small subset of tokens is highly significant in the generation. Scissorhands \cite{NEURIPS2023_a452a7c6} revealed the persistence of importance, indicating that tokens identified as important in initial remain significant throughout subsequent stages of inference. PyramidInfer \cite{yang-etal-2024-pyramidinfer} further explores the distinct attention characteristics across different layers within LLM, and identified that deeper layers exhibit greater redundancy. 

The insights from these studies have inspired us to integrate information from the attention mechanism into our prompt compression methods. Nevertheless, the main objectives of these prior works are distinct from those of our research.
\section{Preliminaries}

\subsection{Problem Formulation}
We first formally define the compression process with target budget. Denote original input tokens as ${\bm{x}} = \{{x}_i\}_{i=1}^{{L}}$ and tokens after compression as $\bm{\widetilde{x}} = \{\widetilde{x}_i\}_{i=1}^{\widetilde{L}}$, where $\bm{\widetilde{x}}$ is a subset of ${\bm{x}}$. The compression rate can be derived from $\tau = \widetilde{L} / L$. Our goal is to find a subset chosen policy that the output of the generative process is comparable to the original prompt, which can be expressed by the formula:
\begin{equation}
\min_{x, \tau} \mathcal{D}(P(\widetilde{y} | \widetilde{x}), P(y | x))
\label{eq:distribute_allign}
\end{equation}
where function D(,) denotes the distance between two distributions such as KL divergence. $\widetilde{y}$ represents LLM generate results from the compressed context $\bm{\widetilde{x}}$, and ${y}$ represents the original output derived from ${\bm{x}}$.

\subsection{Information Entropy}
From an information-theoretic perspective, an effective compression algorithm should strive to minimize the loss of information. The information entropy of the token can be quantified as the output distribution during its autoregressive generation.
This can be expressed as follow:
\begin{equation}
I_t(x) = - \log_2 P(x_t | x_0, x_1, ..., x_{t-1})
\label{eq:information_entropy}
\end{equation}
where $I_t(x)$ represents the information entropy of token $x_t$ and $P(x)$ denotes the output probability while generating token $x_t$. Consequently, a token with a higher certainty in its probability distribution indicates a lower information entropy and thus conveys less information, which can be the guidance during compression.

\subsection{Attention Scores}

The attention mechanism is essential in the Transformer architecture, as it enables the model to focus on critical segments of the input sequence, thereby enhancing its capability to manage long-range dependencies. We denote query matrix as $Q \in \mathbb{R}^{ n \times d}$ and key matrix as $K \in \mathbb{R}^{ n \times d}$ in attention mechanism. Then the normalized attention matrix of the i-th layer and the j-th head can be expressed as $Softmax\left(\frac{Q_{ij} K_{ij}^\top}{\sqrt{d_h}}\right) \in \mathbb{R}^{n \times n}$. Suppose that each element in this matrix is denoted as $q_{uv}$, then the accumulated attention score vector of this matrix can be calculated by:
\begin{equation}
F^{ij}_{score} = (s^{ij}_1, s^{ij}_2, \ldots, s^{ij}_n),  s^{ij}_v = \sum_{u=1}^{n} q_{uv}
\label{eq:attn_score_1}
\end{equation}
where ${s^{ij}_v}$ denotes the accumulated attention score of v-th token in the i-th layer and the j-th head. All accumulated attention score vectors are aggregated into the final one by computing the mean across all layers and heads:
\begin{equation}
\small
\overline{F_{score}} = \frac{1}{MN} \cdot \sum_{i=1}^{N} \sum_{j=1}^{M} F^{ij}_{score} = (\overline{s_1}, \overline{s_2}, \ldots, \overline{s_n})
\label{eq:attn_score_2}
\end{equation}

\section{Methodology}
In this section, we introduce the Dynamic Attention-aware Compression framework, as shown in Figure \ref{fig:illus}. Prior to describing the framework in detail, we first present two key observations that motivated our approach.

\begin{figure*}[!htb]
\hspace{-20pt}
\subfloat{
\includegraphics[scale=0.27,valign=t]{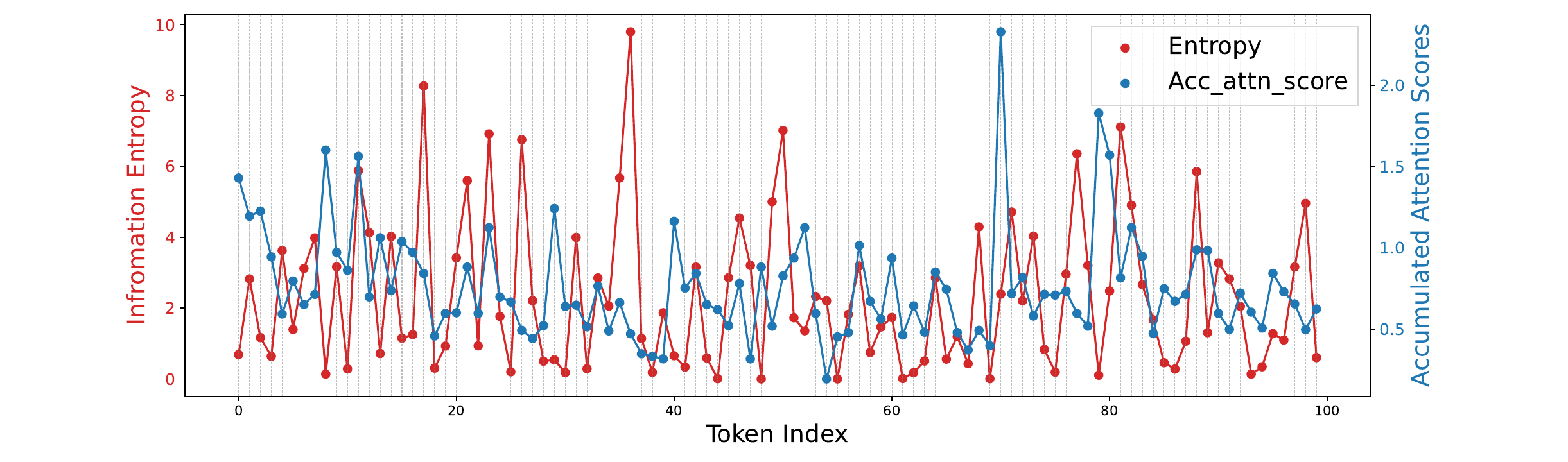}
\label{fig:ob1a}
}
\hspace{-20pt}
\subfloat{
\includegraphics[scale=0.25,valign=t]{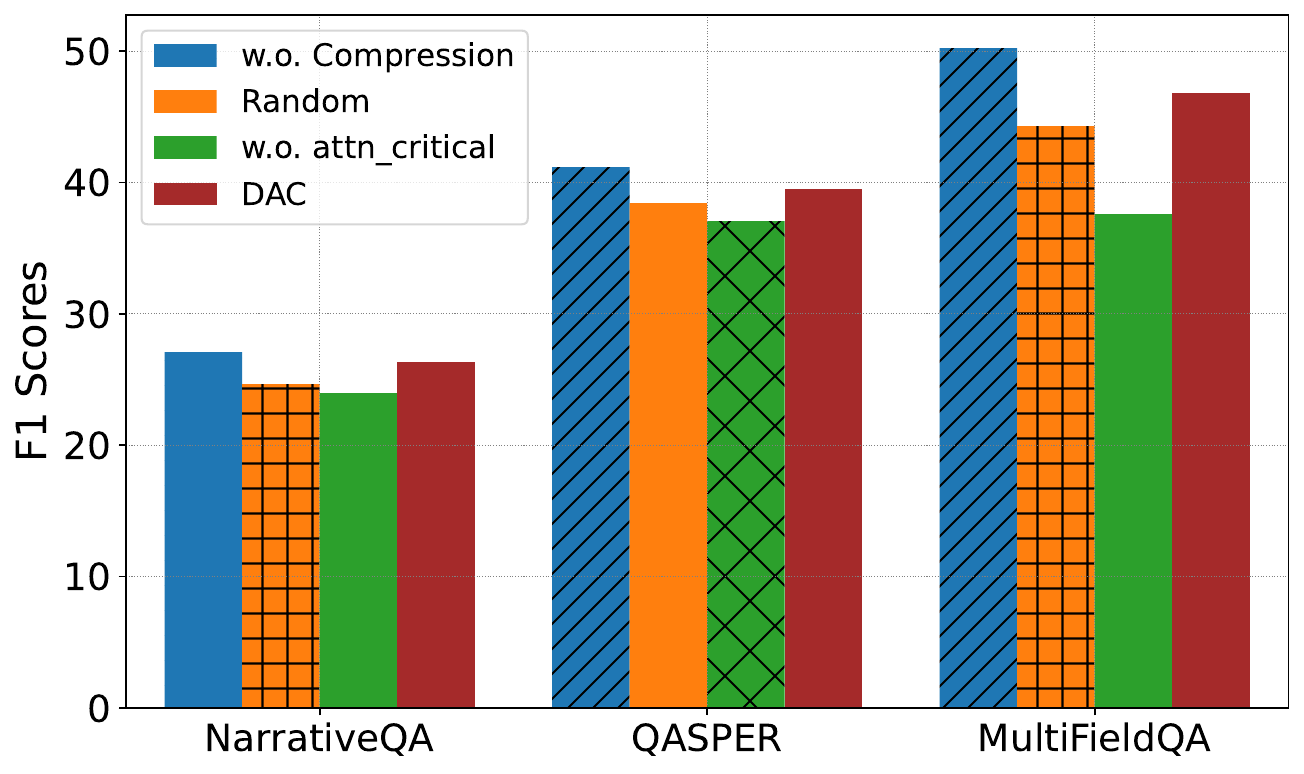}
\label{fig:ob1b}
}
\caption{Left: The information entropy (red) and corresponding accumulated attention scores (blue) on one sequence. It reveals that attention-critical tokens (with high accumulated attention scores) do not necessarily possess high entropy. Right: The performance comparison across four methods: w.o. compression, random, w.o. attention-critical tokens and our DAC method. Without attention-critical tokens, the model's performance on all three datasets significantly deteriorates, even performing worse than random selection methods. Our DAC method effectively addresses this issue by identifying the attention-critical tokens.}
\label{fig:ob1}
\end{figure*}

\begin{figure*}[!htb]
\subfloat{
\includegraphics[scale=0.18,valign=t]{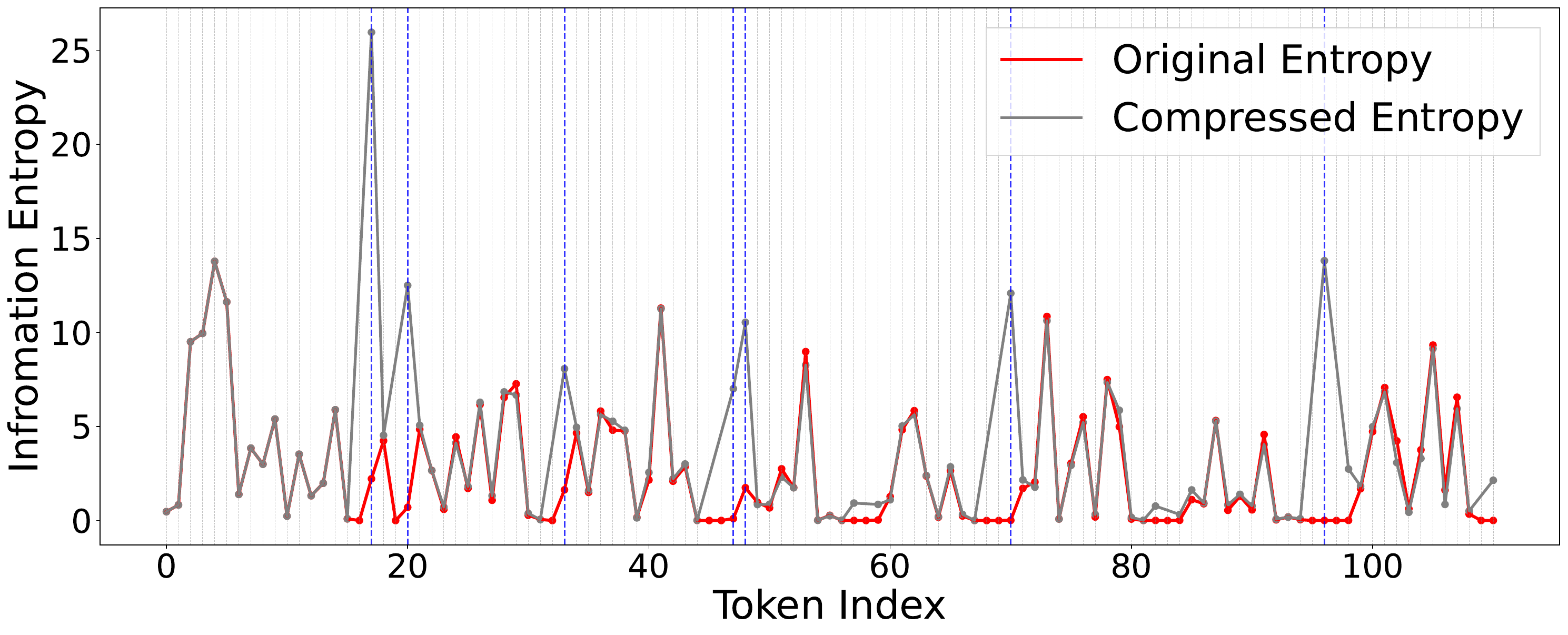}
\label{fig:ob2a}
}
\hspace{10pt}
\subfloat{
\includegraphics[scale=0.25,valign=t]{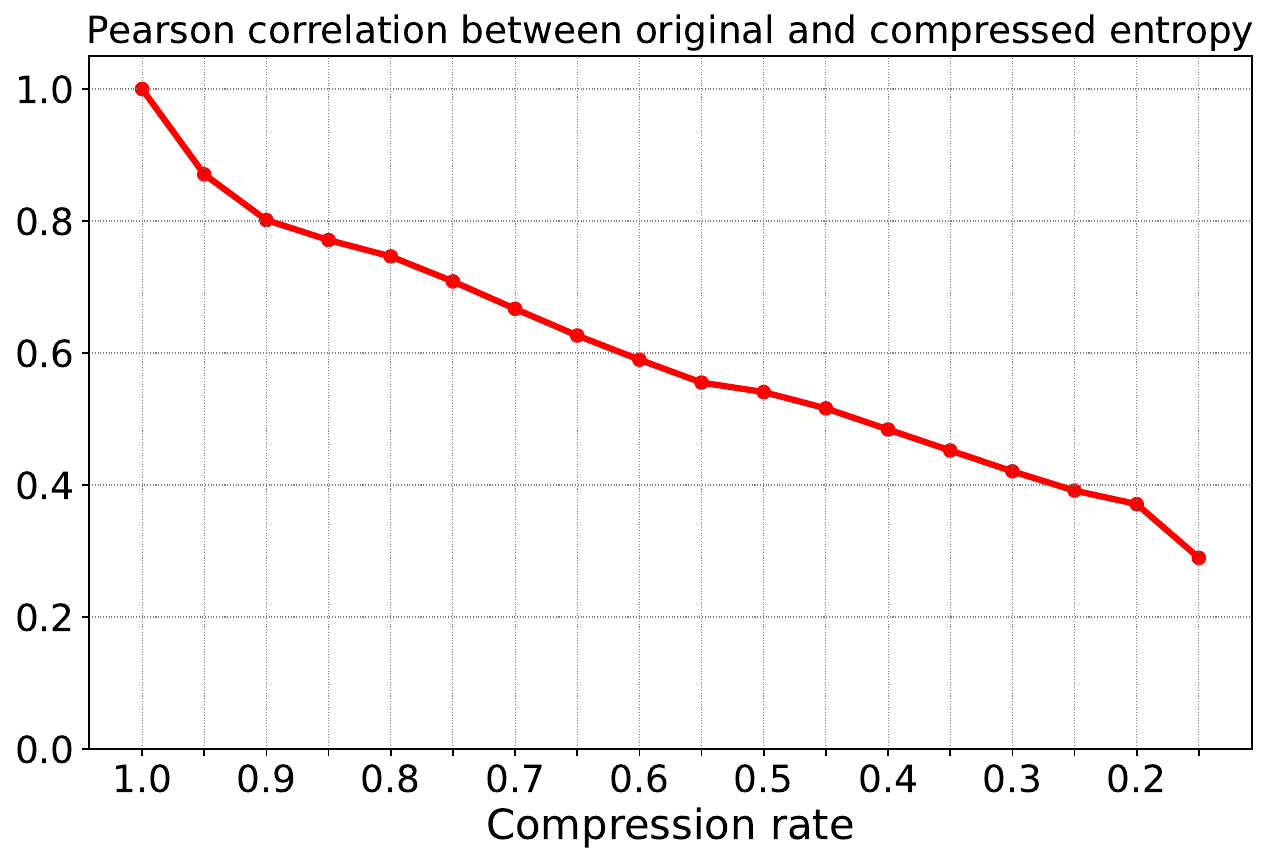}
\label{fig:ob2b}
}
\caption{Left: Entropy differences highlight significant shifts in tokens whose preceding tokens were removed during compression. Right: Compression ratio inversely correlates with entropy differences.}
\label{fig:ob2}
\end{figure*}

\subsection{Observations}
\paragraph{Observation 1: Attention-critical Tokens Matter.}We first observe the relationship between information entropy and accumulated attention score of each token throughout the inference process. In detail, we select Qwen2-7B as the base model, and employ a popular NarrativeQA benchmark. The visualization results on one of sequences are presented in the left of Figure \ref{fig:ob1}. We initially identify tokens with higher accumulated attention scores (e.g., those exceeding a threshold of 1.0) as attention-critical tokens. Our observations reveal that a substantial number of these attention-critical tokens do not exhibit correspondingly high information entropy, suggesting the absence of a linear relationship between the two metrics. Quantitative analysis of 200 sequences from the NarrativeQA dataset also indicates an average Pearson correlation coefficient of 0.095 between the accumulated attention scores and information entropy. This finding implies that relying solely on information entropy as a criterion for compression might result in the exclusion of numerous attention-critical tokens. 

We further analyzed the implications of ignoring attention-critical tokens. We conducted experiments on three single-document QA benchmarks: NarrativeQA, QASPER, and MultiFieldQA and compared four different methods using the F1 score as the evaluation metric.  These methods included: w.o. compression, random select compression, w.o. attention-critical tokens and our DAC method. The compression rate was set to $\tau =0.9 $. The results are shown in the right of Figure \ref{fig:ob1}. Our findings revealed that w.o. attention-critical tokens in prompt significantly degraded the model's answering performance, performing even worse than random select compression. A more effective perceptual compression scheme should integrate both information entropy and accumulated attention scores. The proposed DAC method achieves superior performance by implementing this integrated scheme.

\paragraph{Observation 2: Entropy Shift during Compression.}Information entropy serves as a fundamental metric in past prompt compression methods. Consequently, it is imperative to analyze characteristics of the entropy changes during the compression process. To this end, Figure \ref{fig:ob2} (left) shows the entropy before and after compression with the ratio of 0.9. The figure reveals that even with a relatively low compression ratio, a significant proportion of tokens exhibit substantial shifts in entropy (marked by blue vertical lines). This phenomenon implies that past non-dynamic approach may ignore these critical entropy shifts, thereby compromising the effectiveness of the compression method. It is also noted that, for the majority of these tokens exhibiting substantial shifts, their preceding tokens have been dropped during compression. This observation also inspires an efficient way to address the entropy shift.

Figure \ref{fig:ob2} (right) presents the Pearson correlation between the original and compressed entropy across varying compression ratios. The results demonstrate that as the compression ratio increases, these shifts become more pronounced. This finding illustrates the necessity of developing a dynamic compression method that can adaptively respond to entropy changes, thus enabling a higher compression rate while maintaining LLM performance.

\subsection{Dynamic Attention-aware Compression}
\label{sec:DACmethod}
Based on observation 1, we first propose a metric that integrates both information entropy and accumulated attention scores for prompt compression. We explore two fusion strategies: additive fusion (Eq.(\ref{eq:integration_metric1})) and multiplicative fusion (Eq.(\ref{eq:integration_metric2})). In the additive fusion, a parameter $\alpha$ is introduced to balance the contributions of the information from both sides, and the optimal $\alpha$ is determined through experimentation.

\begin{equation}
M^{a}_t = (1-\alpha) \cdot I_t(x) + \alpha \cdot \overline{s_t}
\label{eq:integration_metric1}
\end{equation}

\begin{equation}
M^{m}_t =  I_t(x) \cdot \overline{s_t}
\label{eq:integration_metric2}
\end{equation}

Based on the insights derived from observation 2, we propose a dynamic method for prompt compression. Specifically, rather than completing the compression in a single step based on the metric, we divide the process into multiple stages. At each stage, we recalculate the information entropy for usage in the current stage to reduce the impact of the significant entropy shifts observed in observation 2.

Additionally, since observation 2 revealed that most tokens with significant entropy shifts have their preceding tokens compressed, we address this issue by limiting the compression of consecutive tokens. This strategy offers two primary benefits: it helps prevent unexpected compression due to the change of entropy within the same compression stage, and it provides a dynamically adjusted compression rate. That is, the compression rate for each round is dynamically adjusted based on the compression pattern from the previous round, expressed as:
\begin{equation}
\triangle\tau = \tau^{1/D} + \triangle P
\label{eq:cal_rate}
\end{equation}
$\triangle\tau$ and $D$ in the equation represents the compression rate in current stage and the dynamic iterations respectively, while $\triangle P $ denotes the percentage of tokens retained in the previous stage due to the limiting of consecutive compression. 

After getting the compression rate in current stage, we first use a
percentile-based filtering approach for compression. In detail, the threshold of integrated metric can be calculated by:
\begin{equation}
T_{\triangle\tau} = np.percentile([M_0], \ldots,[M_n], \triangle\tau)
\label{eq:cal_threshold}
\end{equation}
Next, We filter out those tokens whose fusion metric exceeds the threshold or whose preceding token has already been compressed, and add them to the set $\bm{\widetilde{x}}$ as the compression result in the current stage, denoted as:
\begin{equation}
\bm{\widetilde{x}} = \{\widetilde x_i \mid M(\widetilde x_{i}) \geq T_{\triangle\tau} \lor \widetilde x_{i-1} \notin \bm{\widetilde{x}}\}
\label{eq:compression}
\end{equation}
The compression procedure is completed by performing above steps in multiple stage upon a preset dynamic iterations. The overall procedure of DAC can be referred to Algorithm \ref{alg:Dynamic Attention-aware Prompt Compression}.

\begin{algorithm}[t]
    \small
    \caption{Dynamic Attention-aware Prompt Compression (DAC).} 
    \textbf{Input}: Prompt to compress ${\bm{x}} = \{{x}_i\}_{i=1}^{{L}}$; Target compression rate $\tau$; Dynamic iterations $D$; A small language model $SLM$;
    \begin{algorithmic}[1]
        \State Calculate the accumulated attention score of each token via Eq.(\ref{eq:attn_score_1}) and Eq.(\ref{eq:attn_score_2})
        \For {$i=1, 2, \ldots, D$}
            \State Calculate the compression rate $\triangle\tau$ via Eq.(\ref{eq:cal_rate})
            \State Update the information entropy by $SLM$ and Eq.(\ref{eq:information_entropy})
            \State Calculate the integrated metric $M_i$ for each token via Eq.(\ref{eq:integration_metric1}) or Eq.(\ref{eq:integration_metric2})
            \State Find the $\triangle\tau$-percentile threshold $T_{\triangle\tau}$ via Eq.(\ref{eq:cal_threshold})
            \For {j in range($\bm{\widetilde{x}}$)}
                \If {$M(\widetilde x_{j}) < T_{\triangle\tau}$}
                    \If{$\widetilde{x}_{j-1} \notin \bm{\widetilde{x}}$}
                        \State $P = P + 1$
                    \Else 
                        \State ${\bm{\widetilde{x}}}$.delete($\widetilde x_{j}$)
                    \EndIf
                \EndIf
            \EndFor
            \State Update differential compression ratio $\triangle P$ by $P / \text{len}({\bm{x}})$
        \EndFor
    \end{algorithmic} 
    \textbf{Output}: The compressed prompt $\bm{\widetilde{x}}$.
    \label{alg:Dynamic Attention-aware Prompt Compression}
\end{algorithm}
    
\section{Experiments}
\subsection{Experiments Setup}

\paragraph{Datasets}We comprehensively evaluate the utility preservation of LLM with different prompt compression methods from two different aspects: (i) Contextual Understanding: we utilize four types of tasks from the LongBench \cite{bai2024longbench}: Single-document QA, Multi-document QA, Summarization, and Few-shot Learning. Each task category includes three specific benchmarks. (ii) Reasoning and In-context Learning: We employ GSM8K \cite{cobbe2021training} and Big Bench Hard (BBH) \cite{suzgun-etal-2023-challenging} datasets. Consistent with prior studies, we adopt Exact Match (EM) as evaluation metric.

\paragraph{Baselines}We take three most effective prior studies as baselines:

\noindent\textbf{Selective-Context} \cite{li-etal-2023-compressing}: Selective-Context employs the information entropy of the most effective lexical units (phrases) as the mertic for compression.

\noindent\textbf{LLMLingua} \cite{jiang-etal-2023-llmlingua}: LLMLingua first involves a budget controller that assigns different compression strategies to various components of the prompt (instruction, demonstrations, and the question). Subsequently, it performs iterative token-level prompt compression, by which the prompt is segmented, and each segment is compressed sequentially based on the information entropy of the tokens.

\noindent\textbf{LLMLingua2} \cite{pan-etal-2024-llmlingua}: LLMLingua2 first distills compressed data from GPT4 using MeetingBank dataset. It then trains a specialize small model for prompt compression based on a transformer encoder architecture.

To ensure a fair comparison, all compression rates in experiments are actual compression rates. For those methods that dynamically adjust the compression rate based on the input prompt, we adjust the target compression rate to ensure that the final number of compressed tokens is approximately consistent across different methods.

\begin{table*}[]
\centering
    \setlength{\tabcolsep}{1mm}
    \renewcommand{\arraystretch}{1.5}
    \vspace{-2ex}
    \resizebox{2.1\columnwidth}{!}{
\begin{tabular}{lccccccccccccccccc}
\toprule
\multicolumn{1}{c|}{} &
  \multicolumn{17}{c}{\textbf{LongBench}} \\ \cline{2-18} 
\multicolumn{1}{c|}{} &
  \multicolumn{4}{c|}{Single-Doc QA} &
  \multicolumn{4}{c|}{Multi-Doc QA} &
  \multicolumn{4}{c|}{Summarization} &
  \multicolumn{4}{c|}{Few-shot Learning} &
   \\
\multicolumn{1}{c|}{\multirow{-3}{*}{\textbf{Methods}}} &
  Nar.QA &
  Qasper &
  Mul.QA &
  \multicolumn{1}{l|}{AVG} &
  Hot.QA &
  2Wi.QA &
  Musique &
  \multicolumn{1}{l|}{AVG} &
  GovRe. &
  QMSum &
  M.News &
  \multicolumn{1}{l|}{AVG} &
  TREC &
  Tri.QA &
  SAMSum &
  \multicolumn{1}{l|}{AVG} &
  \multirow{-2}{*}{\begin{tabular}[c]{@{}c@{}}All\\ AVG\end{tabular}} \\ \hline
\multicolumn{18}{c}{\textit{Compression rate $\tau$ = 0.5}} \\ \hline
\multicolumn{1}{l|}{Selective-Context} &
  18.75 &
  35.07 &
  28.90 &
  \multicolumn{1}{c|}{27.57} &
  38.30 &
  36.07 &
  21.28 &
  \multicolumn{1}{c|}{31.88} &
  25.83 &
  24.11 &
  24.77 &
  \multicolumn{1}{c|}{24.90} &
  29 &
  81.80 &
  38.07 &
  \multicolumn{1}{c|}{49.62} &
  33.50 \\
\multicolumn{1}{l|}{LLMLingua} &
  20.36 &
  28.26 &
  27.69 &
  \multicolumn{1}{c|}{25.44} &
  40.11 &
  35.69 &
  21.00 &
  \multicolumn{1}{c|}{32.27} &
  26.19 &
  23.5 &
  24.71 &
  \multicolumn{1}{c|}{24.80} &
  40 &
  78.07 &
  39.15 &
  \multicolumn{1}{c|}{52.41} &
  33.73 \\
\multicolumn{1}{l|}{LLMLingua-2} &
  24.25 &
  35.22 &
  38.70 &
  \multicolumn{1}{c|}{32.72} &
  43.61 &
  38.11 &
  26.80 &
  \multicolumn{1}{c|}{\textbf{36.17}} &
  26.15 &
  25.54 &
  25.78 &
  \multicolumn{1}{c|}{25.82} &
  35 &
  77.6 &
  40.37 &
  \multicolumn{1}{c|}{50.99} &
  36.43 \\
\multicolumn{1}{l|}{DAC} &
  24.85 &
  33.46 &
  40.12 &
  \multicolumn{1}{c|}{\textbf{32.81}} &
  42.37 &
  39.57 &
  22.44 &
  \multicolumn{1}{c|}{34.79} &
  30.4 &
  25.95 &
  25.77 &
  \multicolumn{1}{c|}{\textbf{27.37}} &
  50 &
  80.00 &
  38.14 &
  \multicolumn{1}{c|}{\textbf{56.05}} &
  \textbf{37.76} \\ \hline
\multicolumn{18}{c}{\textit{Compression rate $\tau$ = 0.2}} \\ \hline
\multicolumn{1}{l|}{Selective-Context} &
  16.83 &
  29.25 &
  25.57 &
  \multicolumn{1}{c|}{23.88} &
  36.58 &
  33.09 &
  18.08 &
  \multicolumn{1}{c|}{29.25} &
  22.3 &
  23.57 &
  21.77 &
  \multicolumn{1}{c|}{22.55} &
  21.5 &
  75.92 &
  37.67 &
  \multicolumn{1}{c|}{45.03} &
  30.18 \\
\multicolumn{1}{l|}{LLMLingua} &
  18.35 &
  21.78 &
  24.65 &
  \multicolumn{1}{c|}{21.59} &
  38.43 &
  32.85 &
  21.95 &
  \multicolumn{1}{c|}{31.08} &
  22.91 &
  22.38 &
  22.5 &
  \multicolumn{1}{c|}{22.60} &
  30 &
  77.83 &
  36.07 &
  \multicolumn{1}{c|}{47.97} &
  30.81 \\
\multicolumn{1}{l|}{LLMLingua-2} &
  19.47 &
  30.45 &
  25.86 &
  \multicolumn{1}{c|}{25.26} &
  40.26 &
  33.32 &
  21.85 &
  \multicolumn{1}{c|}{\textbf{31.81}} &
  21.56 &
  24.32 &
  22.48 &
  \multicolumn{1}{c|}{22.79} &
  24 &
  78.92 &
  33.23 &
  \multicolumn{1}{c|}{45.38} &
  31.31 \\
\multicolumn{1}{l|}{DAC} &
  18.16 &
  27.84 &
  30.18 &
  \multicolumn{1}{c|}{\textbf{25.39}} &
  38.68 &
  31.72 &
  22.49 &
  \multicolumn{1}{c|}{30.96} &
  25.73 &
  23.32 &
  23.59 &
  \multicolumn{1}{c|}{\textbf{24.21}} &
  31 &
  80.41 &
  38.47 &
  \multicolumn{1}{c|}{\textbf{49.96}} &
  \textbf{32.63} \\ \hline \hline
\multicolumn{1}{l|}{Original Prompt} &
  {\color[HTML]{333333} 27.08} &
  41.1 &
  50.23 &
  \multicolumn{1}{c|}{39.47} &
  56.92 &
  55.65 &
  36.56 &
  \multicolumn{1}{c|}{49.71} &
  35.2 &
  26.02 &
  24.35 &
  \multicolumn{1}{c|}{28.52} &
  78 &
  87.19 &
  45.03 &
  \multicolumn{1}{c|}{70.07} &
  46.94 \\
\multicolumn{1}{l|}{Zero-Shot} &
  14.92 &
  18.09 &
  19.11 &
  \multicolumn{1}{c|}{17.37} &
  17.11 &
  25.88 &
  6.35 &
  \multicolumn{1}{c|}{16.45} &
  19.92 &
  10.27 &
  8.55 &
  \multicolumn{1}{c|}{12.91} &
  3 &
  73.43 &
  30.89 &
  \multicolumn{1}{c|}{21.59} &
  17.08 \\ \bottomrule
\end{tabular}}
\caption{Performance of different methods under different compression rates on the LongBench (Qwen2-7B). The results of original prompt and zero-shot experiments are also shown at the bottom.}
\label{tab:longbench}
\end{table*}

\paragraph{Implementation Details}
Our experiments are conducted on two kinds of LLMs. The main results are performed on the Qwen2 series models\footnote{https://github.com/QwenLM/Qwen}. Specifically, the SLM used for compression is Qwen2-0.5B, and the LLM used for evaluation is Qwen2-7B. Additionally, to study DAC's adaptation to various LLMs, we also use the LLaMA3 series models\footnote{https://github.com/meta-llama/llama3} to measure the impact of different models (see Section \ref{diff_model}). For these experiments, the SLM is LLaMA 3.2-1B, and the LLM is LLaMA 3.1-8B. The Dynamic times $D$ in algorithm is set by $D = L_{input}/100$, with the maximum value of 15.
All experiments are conducted on an NVIDIA A800 GPU.  We use greedy decoding to ensure the stability of the experimental results. The experimental environment includes the following configurations: CUDA Version 12.0, PyTorch version 2.4.0, and HuggingFace's Transformers\footnote{https://github.com/huggingface/transformers} with version 4.45.1.

\subsection{Fusion Strategies}
\begin{figure}[!htb]
\hspace{10pt}
\includegraphics[scale=0.3]{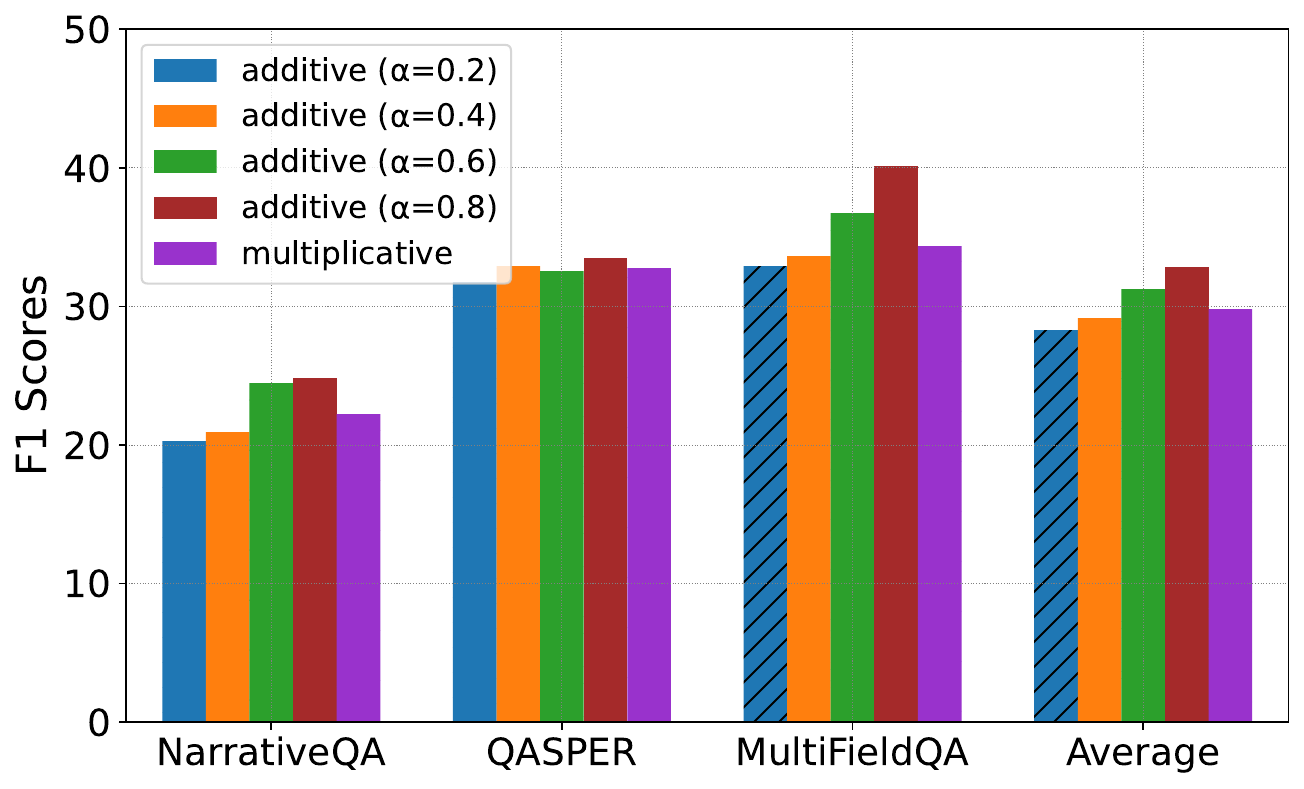}
\caption{The performance of DAC method with different fusion strategies.}
\label{fig:para_select}
\end{figure}

In section \ref{sec:DACmethod}, we discussed two fusion strategies for integrating information entropy and accumulated attention scores: additive fusion with parameter $\alpha$ and multiplicative fusion. In this section, we conduct experiments to find the appropriate fusion strategy. We consider five situations: additive with $\alpha=0.2, 0.4, 0.6, 0.8$ and multiplicative. The experiments are conducted on single-document QA task from the Longbench, and the results are shown in Figure \ref{fig:para_select}. The results show that the best performance is achieved with additive fusion with  $\alpha=0.8$.  Therefore, subsequent experiments will adopt this fusion strategy.

\subsection{Main Results}
\label{sec:main_res}
Table \ref{tab:longbench} shows the comparative performance of proposed DAC method against the baselines on Longbench with two compression rates ($\tau=0.2$ and $\tau=0.5$ ). Overall, DAC method outperforms the baselines in the task types of single-document QA, summarization, and few-shot learning, as well as in the overall average score.

Table \ref{tab:gsmbbh} presents the comparative results on the GSM8K and BBH datasets. It can be seen that the DAC method achieves the best performance in most cases. With the compression rate $\tau=0.5$, the DAC method's performance on GSM8K decreases by only 0.84 compared to the original prompt, demonstrating its performance retention in reasoning tasks. We also notice that although the DAC method is behind LLMLingua by 0.67 on the BBH dataset with $\tau=0.5$,  it outperforms LLMLingua by 0.8 with $\tau=0.2$. This demonstrates that the introduction of the dynamic mechanism of information entropy enables DAC to perform better with high compression rates.

The experimental results also suggest that compression methods based on information entropy may be more generalized. Training specialized small models for compression, such as LLMLingua2, may suffer from reduced generalization due to limitations in the training datasets. For instance, while LLMLingua2 performs very well on LongBench, especially for multi-document QA tasks, it falls short compared to entropy-based method on GSM8K and BBH datasets.

\subsection{Ablation Study}

To validate the effectiveness of each component in the proposed DAC method, we conducted ablation studies with three different configurations: \textit{DAC w/o Attention-aware Metric} indicates that the accumulated attention score is not involved, and only information entropy is used as the metric for compression. \textit{DAC w/o Dynamic Procedure} indicates that the dynamical observation of entropy shift is not used, and the entropy is calculated once in the beginning instead. \textit{DAC w/o Limiting Consecutive Compression} indicates that consecutive tokens are allowed to be compressed in same stage of dynamic procedure. The results are shown in Table \ref{tab:ablation}.

\begin{table}[]
\centering
\footnotesize
\setlength{\tabcolsep}{1mm}
\begin{tabular}{lcc}
\toprule
\multicolumn{1}{l|}{\textbf{Methods}}           & \textbf{GSM8K}          & \textbf{BBH}            \\ \hline
\multicolumn{3}{c}{\textit{Compression rate $\tau$ = 0.5}}                        \\ \hline
\multicolumn{1}{l|}{Selective-Context} & 61.33          & 50.07          \\
\multicolumn{1}{l|}{LLMLingua}         & 72.86          & \textbf{54.98} \\
\multicolumn{1}{l|}{LLMLingua-2}        & 67.85          & 47.74          \\
\multicolumn{1}{l|}{DAC}               & \textbf{74.37} & 54.31          \\ \hline
\multicolumn{3}{c}{\textit{Compression rate $\tau$ = 0.2}}                        \\ \hline
\multicolumn{1}{l|}{Selective-Context} & 60.12          & 46.66          \\
\multicolumn{1}{l|}{LLMLingua}         & 65.73          & 49.61          \\
\multicolumn{1}{l|}{LLMLingua-2}        & 66.49          & 44.16          \\
\multicolumn{1}{l|}{DAC}               & \textbf{67.85} & \textbf{50.41} \\ \hline \hline
\multicolumn{1}{l|}{Original Prompt}   & 75.21          & 60.70          \\
\multicolumn{1}{l|}{Zero-Shot}         & 42.61          & 37.75          \\ \bottomrule
\end{tabular}
\caption{Performance of different methods under different compression rates on the GSM8K and BBH dataset (Qwen2-7B).}
\label{tab:gsmbbh}
\end{table}

First, it can be observed that the F1 score of Single-Doc QA drops the most when DAC is w/o Attention-aware Metric. This significant decline occurs when only information entropy is used, without considering attention between tokens. This is consistent with our observation 1, which states that not only should information entropy be used as a guideline for compression, but it is also crucial to retain tokens that are important in the attention mechanism. A metric that can integrate both high-level information entropy and low-level information in algorithm itself can significantly enhance the performance. Then compared with DAC w/o Dynamic Procedure, it reveals that the introduced dynamic procedure can identify essential information during the compression process, which would be missed without dynamic detection of entropy. Finally, it also can be found that DAC w/o Limiting Consecutive Compression will degrade the performance slightly. We conjecture that this could be due to the inappropriate compression of a subsequent token, which was caused by the dropping of its preceding token and then resulting entropy shift.

\begin{table}[]
\resizebox{1\columnwidth}{!}{
\begin{tabular}{lc}
\toprule
                                     & Single-Doc QA \\ \midrule
DAC                                  & 32.81         \\
- w/o Attention-aware Metric           & 28.16         \\
- w/o Dynamic Procedure                & 29.84         \\
- w/o Limiting Consecutive Compression & 31.88         \\ \bottomrule
\end{tabular}}
\caption{Ablation study on single-document QA with compression rates $\tau=0.5$.}
\label{tab:ablation}
\end{table}

\subsection{Different Models}
\label{diff_model}
To ensure the effectiveness of our method across different model types, we conducted experiments on other models. Specifically, here we use LLaMA 3.2-1B as the SLM for compression and LLaMA 3.1-8B as the LLM for evaluation. The experimental results are shown in Table \ref{tab:llama3_res}. For simplicity, we report the average scores across different task types. The experimental results are shown in Table 4. It can be observed that DAC also demonstrates excellent performance on LLaMA3 series models, achieving state-of-the-art results in most task types and in the overall average score.

\begin{table}[]
\centering
\resizebox{1\columnwidth}{!}{
\begin{tabular}{llllll}
\toprule
\multicolumn{1}{c}{\multirow{2}{*}{Methods}} &
  \multicolumn{4}{c}{LongBench} &
  \multicolumn{1}{c}{\multirow{2}{*}{\begin{tabular}[c]{@{}c@{}}ALL\\ AVG\end{tabular}}} \\ 
\multicolumn{1}{c}{} &
  \multicolumn{1}{c}{Sin.QA} &
  \multicolumn{1}{c}{Mul.QA} &
  \multicolumn{1}{c}{Summ.} &
  \multicolumn{1}{c}{Few.} &
  \multicolumn{1}{c}{} \\ \midrule
Selective-Context & 26.40          & 24.78          & 21.98          & 55.47          & 32.16          \\
LLMLingua         & 27.81          & 24.68          & \textbf{23.13} & 57.63          & 33.31          \\
LLMLingua-2       & 32.51          & 27.93          & 22.74          & 56.08          & 34.82          \\
DAC               & \textbf{32.68} & \textbf{28.08} & 22.92          & \textbf{58.01} & \textbf{35.42} \\ \midrule
Original Prompt   & 37.35          & 36.36          & 27.39          & 71.01          & 43.03          \\
Zero-Shot         & 14.83          & 17.23          & 13.61          & 42.29          & 21.99          \\ \bottomrule
\end{tabular}}
\caption{The comparision results on LongBench using the LLaMA3 series models (compression rates $\tau=0.5$)}
\label{tab:llama3_res}
\end{table}

\subsection{Overhead Analysis}

We analyzed the overhead introduced by compression using different methods. Specifically, the profilling of overhead is conducted on a random sample from the GovReport benchmark, which contains 12,908 tokens of prompt. The length of the output tokens is set to 500, and the compression rate is 0.2. We followed the experimental setup for section \ref{sec:main_res}, where the SLM is Qwen2-0.5B and the LLM is Qwen2-7B.

\begin{figure}[!htb]
\centering
\includegraphics[scale=0.4]{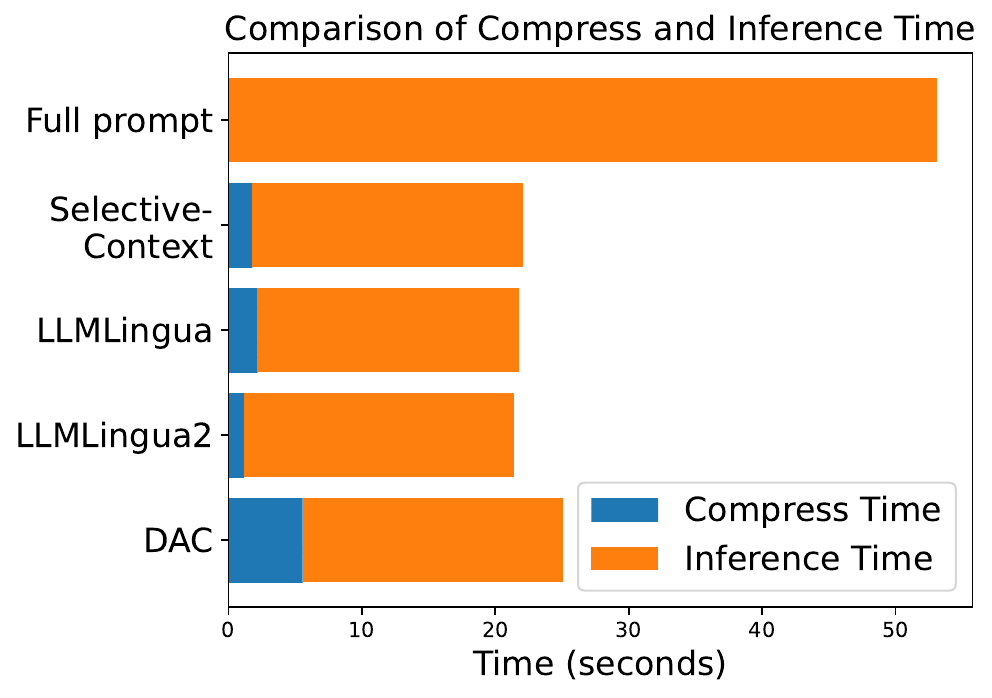}
\caption{The comparison of compress and inference time using different methods}
\label{fig:overhead}
\end{figure}

For each compression method, we plot the time taken for compression and the actual inference time. We also record the time required for using the full prompt. The results are shown in Figure \ref{fig:overhead}. As can be seen, the DAC method, due to the introduction of the dynamic procedure, takes longer time in compression compared to other methods, as indicated by blue in the bar chart. However, this additional time is quite small when compared to the time saved by compression (compared to the full prompt). Additionally, it is worth noting that the parameter of LLM in production is often much larger than 7B, which further amplify the benefits of compression in terms of reduced inference time and memory.

\subsection{More Compression Rates}

\begin{figure}[!htb]
\centering
\includegraphics[scale=0.3]{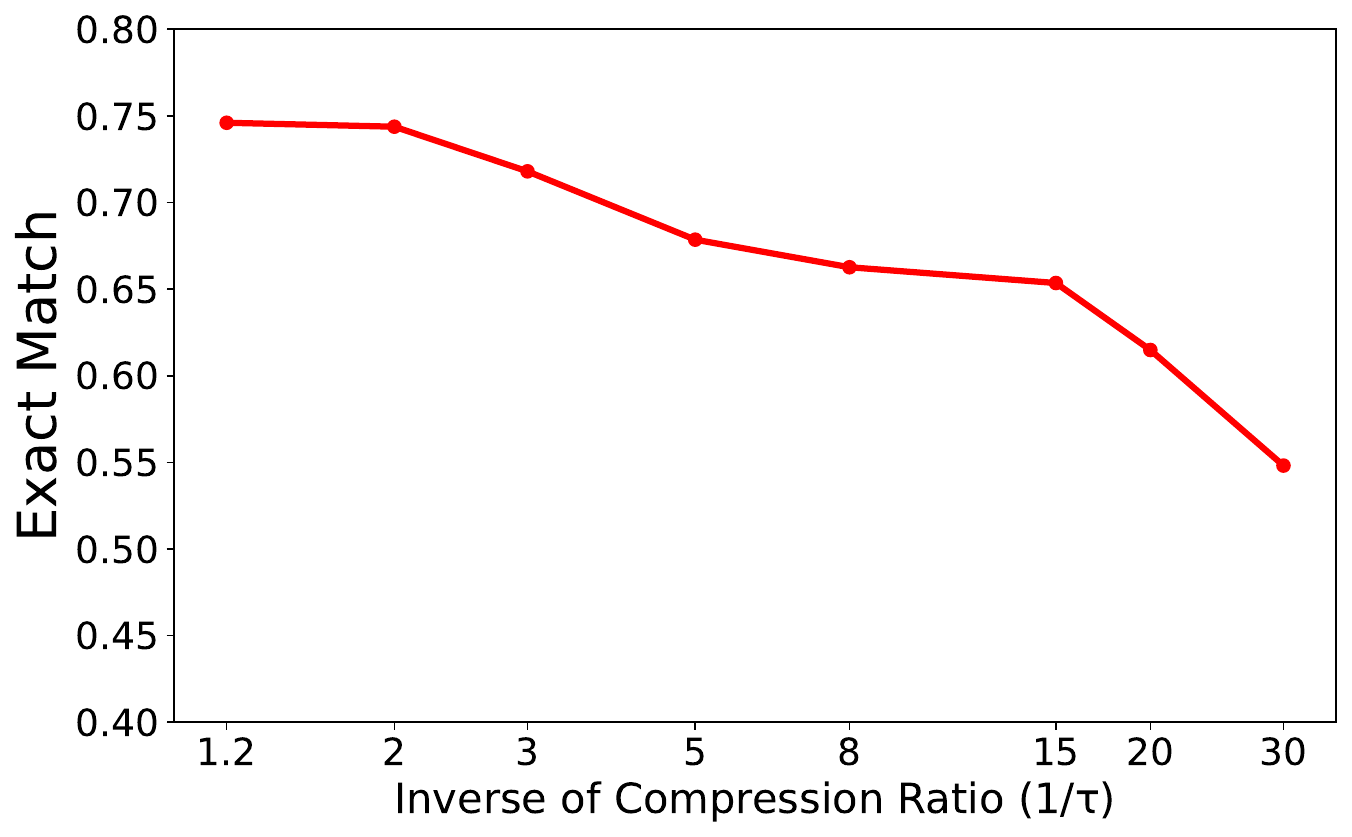}
\caption{The performance of DAC at various compression rates on GSM8K (Qwen2-7B).}
\label{fig:cr}
\end{figure}

To assist users in achieving a balanced trade-off between utility preserving and efficiency, we present a detailed analysis of how various compression rates affect the performance of our DAC method. The experiments are conducted on the GSM8K dataset, and the results are shown in the Figure \ref{fig:cr}. It can be observed that when the compression rate exceeds 0.5 ($1/\tau < 2$), the model's utility preservation is relatively good. For scenarios that prioritize low compression rates and can tolerate a certain degree of performance degradation, setting $1/\tau$ to around 8 offers a feasible solution.

\section{Conclusion}
This paper addresses task-agnostic prompt compression by proposing a dynamic attention-aware method. This approach aims to overcome the limitations identified in existing work, achieving fine-grained compression through the integration of information across different levels. We conduct extensive experiments and analyses across various domains, including contextual understanding, reasoning, and in-context learning. Our approach outperforms strong baselines across various domains and different series of LLMs, while introducing only acceptable additional overhead. The results indicate significant practical implications of our method for enabling LLMs to save computational costs and handle longer contexts effectively.

\section*{Limitations}
There are also some limitations in our approach: (1) The current implementation of DAC requires obtaining attention matrices from all layers and heads, which necessitates the development of a method to identify the most representative attention matrices for more efficient information fusion. Additionally, DAC is not compatible with high-efficiency attention methods (e.g., Flash Attention) as it does not require calculating attention scores. The application of DAC on such methods will result in additional attention score calculations. (2) While the existing dynamic procedure supports adjusting the number of dynamic iterations based on context length, it hits an upper limit when the context becomes excessively long. This can lead to performance degradation as the granularity of perception becomes coarser. A potential solution could involve developing a method that senses information density to adaptively adjust the number of dynamic iterations, thereby maintaining performance even with very long contexts.

\bibliography{custom}

\clearpage

\appendix

\section{Why Small Models Can Be Used for Compression}.

Previous works have typically employed a small language model for compression but has not extensively discussed the underlying reasons. In this section, we provide empirical evidence to support why the entropy from a smaller model can effectively aid a larger model in identifying salient information. 

\begin{figure*}[!b]
\centering
\includegraphics[scale=0.45]{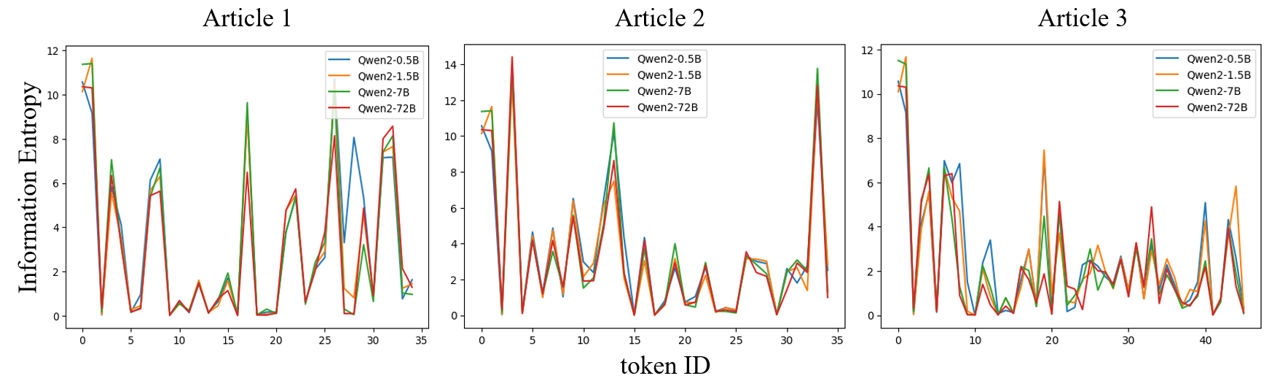}
\caption{Visualization of the entropy similarities between different model parameter sizes (Qwen2 series
models).}
\label{fig:entropy_vir}
\end{figure*}

\begin{figure}[!htb]
\centering
\includegraphics[scale=0.7]{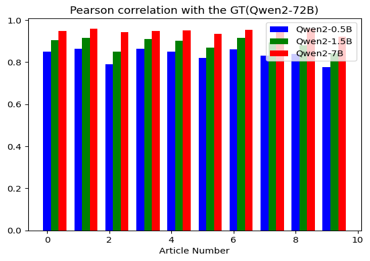}
\caption{Quantitative analysis of the entropy similarities between different model parameter sizes (Qwen2 series
models).}
\label{fig:entropy_sim}
\end{figure}

We conducted experiments on the ArXiv corpus \cite{li-etal-2023-compressing} using various sizes of the Qwen2 models, ranging from 0.5B to 72B parameters, to analyze their entropy outputs. For clearer visualization, we focused on the initial tokens, as illustrated in Figure \ref{fig:entropy_vir}. Despite substantial differences in model size, all models exhibit remarkable consistency in their entropy across different texts. Figure \ref{fig:entropy_sim} presents the quantitative results of entropy similarities using Pearson correlation. Here, we consider the entropy output from Qwen2-72B as the ground truth (GT) and compare it with the entropy outputs from smaller models. The x-axis represents different articles. The data show that while the similarity to GT increases with model size, even the smallest 0.5B model achieves an average similarity of 0.835. The consistent entropy patterns across different model sizes demonstrate that the critical information captured by large models can be efficiently approximated by smaller models.



\section{Compression Case Study}

\begin{figure*}[!htb]
\centering
\includegraphics[scale=0.75]{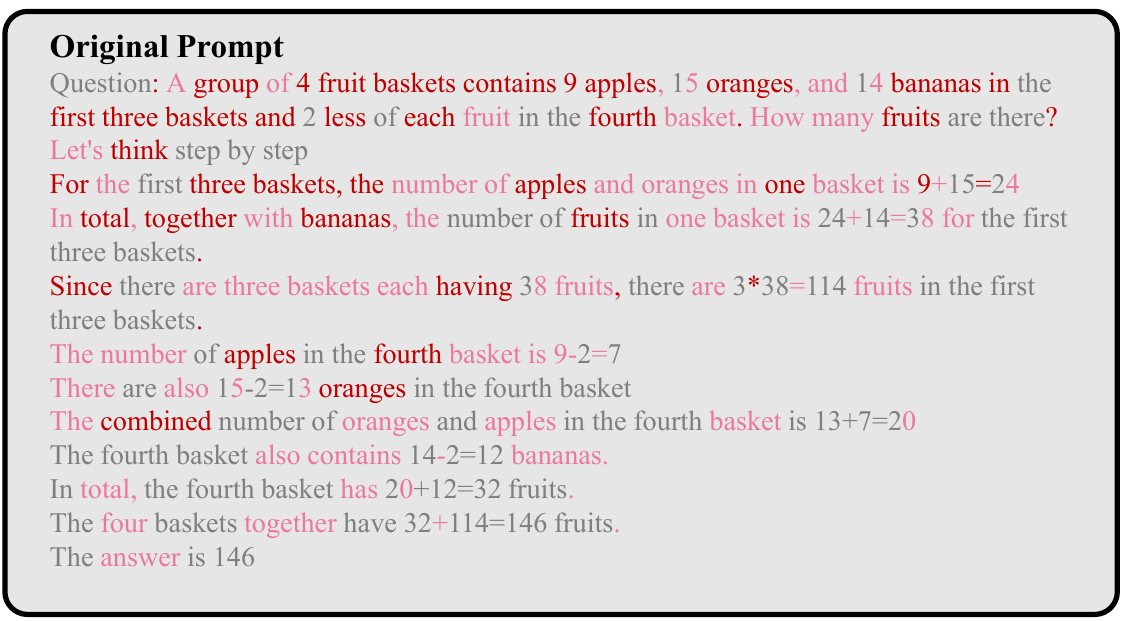}
\caption{Cases study on GSM8K dataset.}
\label{fig:GSM8K_prompt}
\end{figure*}

\begin{figure*}[!htb]
\centering
\includegraphics[scale=0.82]{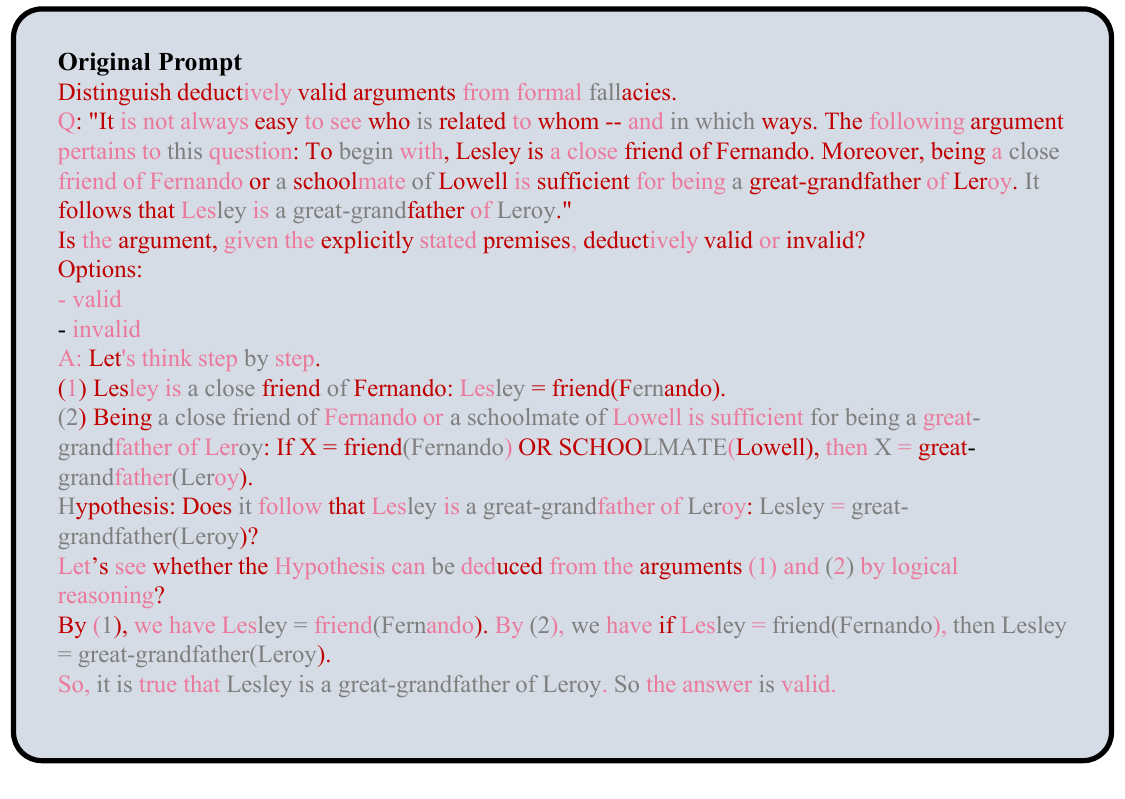}
\caption{Cases study on formal\_fallacies of BBH dataset.}
\label{fig:BBH_prompt}
\end{figure*}

We present various compresstion examples in Figure \ref{fig:GSM8K_prompt} and Figure \ref{fig:BBH_prompt} using DAC. In each example, tokens preserved at a compression rate of 0.2 are highlighted in dark red, while those preserved at a compression rate of 0.5 are shown in light red.

\end{document}